\documentclass[journal]{IEEEtran}
\usepackage{graphicx}
\usepackage{booktabs}
\usepackage{multirow}
\usepackage{xcolor}
\usepackage{array}
\usepackage{amsmath}
\usepackage[colorlinks=true,linkcolor=blue]{hyperref}
\begin{document}

\title{Cross-Modal Object Tracking via Modality-Aware Fusion Network and A Large-Scale Dataset}
\author{Lei Liu, Mengya Zhang, Cheng Li, Chenglong Li, and Jin Tang
\thanks{This research is partly supported by the National Natural Science Foundation of China (No. 62376004), the Natural Science Foundation of Anhui Province (No. 2208085J18), and the Natural Science Foundation of Anhui Higher Education Institution (No. 2022AH040014). (Corresponding author: Chenglong Li.)}
\thanks{Chenglong Li is affiliated with the Information Materials and Intelligent Sensing Laboratory of Anhui Province, Anhui Provincial Key Laboratory of Multimodal Cognitive Computation, School of Artificial Intelligence, Anhui University, Hefei 230601, China (e-mail: lcl1314@foxmail.com).}
\thanks{Lei Liu, Mengya Zhang, and Jin Tang are affiliated with the Information Materials and Intelligent Sensing Laboratory of Anhui Province, Key Laboratory of Intelligent Computing and Signal Processing of the Ministry of Education, Anhui Provincial Key Laboratory of Multimodal Cognitive Computation, School of Computer Science and Technology, Anhui University, Hefei 230601, China (e-mail: liulei970507@163.com; 2846190720@qq.com; tangjin@ahu.edu.cn).}
\thanks{Cheng Li is affiliated with the School of Computer Science and Technology, Anhui University, Hefei 230601, China (e-mail: farawaylc@qq.com).}
}
% The paper headers
\markboth{Journal of \LaTeX\ Class Files,~Vol.~14, No.~8, August~2015}%
{Shell \MakeLowercase{\textit{et al.}}: Bare Demo of IEEEtran.cls for IEEE Journals}

\maketitle
\begin{abstract}
Visual tracking often faces challenges such as invalid targets and decreased performance in low-light conditions when relying solely on RGB image sequences. While incorporating additional modalities like depth and infrared data has proven effective, existing multi-modal imaging platforms are complex and lack real-world applicability. In contrast, near-infrared (NIR) imaging, commonly used in surveillance cameras, can switch between RGB and NIR based on light intensity. However, tracking objects across these heterogeneous modalities poses significant challenges, particularly due to the absence of modality switch signals during tracking.
To address these challenges, we propose an adaptive cross-modal object tracking algorithm called \textbf{M}odality-\textbf{A}ware \textbf{F}usion \textbf{Net}work (\textbf{MAFNet}). 
MAFNet efficiently integrates information from both RGB and NIR modalities using an adaptive weighting mechanism, effectively bridging the appearance gap and enabling a modality-aware target representation. It consists of two key components: an adaptive weighting module and a modality-specific representation module. The adaptive weighting module predicts fusion weights to dynamically adjust the contribution of each modality, while the modality-specific representation module captures discriminative features specific to RGB and NIR modalities. MAFNet offers great flexibility as it can effortlessly integrate into diverse tracking frameworks. With its simplicity, effectiveness, and efficiency, MAFNet outperforms state-of-the-art methods in cross-modal object tracking.
To validate the effectiveness of our algorithm and overcome the scarcity of data in this field, we introduce CMOTB, a comprehensive and extensive benchmark dataset for cross-modal object tracking. CMOTB consists of 61 categories and 1000 video sequences, comprising a total of over 799K frames.
We believe that our proposed method and dataset offer a strong foundation for advancing cross-modal object tracking research. The dataset, toolkit, and source code will be publicly available at: \href{https://github.com/mmic-lcl/Datasets-and-benchmark-code}{\underline{{https://github.com/mmic-lcl/Datasets-and-benchmark-code}}}.
\end{abstract}


\IEEEpeerreviewmaketitle

\section{Introduction}
\IEEEPARstart{V}{isual} object tracking is a fundamental task in computer vision, involving locating a target object in subsequent video frames given its initial position in the first frame. It plays a crucial role in various visual systems, such as video surveillance%\cite{mishra2016study}
, intelligent transportation%\cite{liu2023efficient}
, and human-computer interaction%\cite{bozomitu2019development}
% , where accurate localization and tracking of objects are essential
. However, existing tracking methods often rely on RGB image sequences, which are sensitive to changes in lighting conditions\cite{huang2019got, zhu2023tiny}
%\cite{wu2013online, huang2019got, zhu2023tiny}
. As a result, these methods struggle to locate targets effectively in low-light conditions, leading to a significant degradation in tracking performance. Therefore, there is an urgent need to develop trackers that can maintain robust performance even in varying lighting conditions.

To overcome the limitations of the visible light modality, a viable solution is to incorporate other modalities that are less affected by lighting variations. These modalities can include thermal infrared modality that perceives object thermal radiation information\cite{li2020challenge, zhang2022visible, lu2022duality, li2021rgbt}, depth modality that perceives distance information\cite{yan2021depthtrack, song2013tracking}, or event modality that perceives motion information\cite{wang2023visevent}. While incorporating these multi-modal data can effectively overcome challenges posed by lighting variations, it also introduces new problems. For instance, the thermal infrared modality has lower resolution and is susceptible to interference from background temperature. The depth modality is generally more suitable for indoor scenes and has a limited sensing range. The event modality is not sensitive to slow-moving objects and exhibits lower frame rates in low-light environments. Additionally, these modalities have imaging characteristics that significantly differ from the visible light modality, often requiring independent sensors. Moreover, precise multi-modal data alignment is essential to effectively utilize the multi-modal information. However, this process introduces additional time and resource consumption and the need for complex multi-modal imaging platform design \cite{li2021lasher}.

\begin{figure*}[!htbp]
	\centering
    \includegraphics[width=\linewidth]{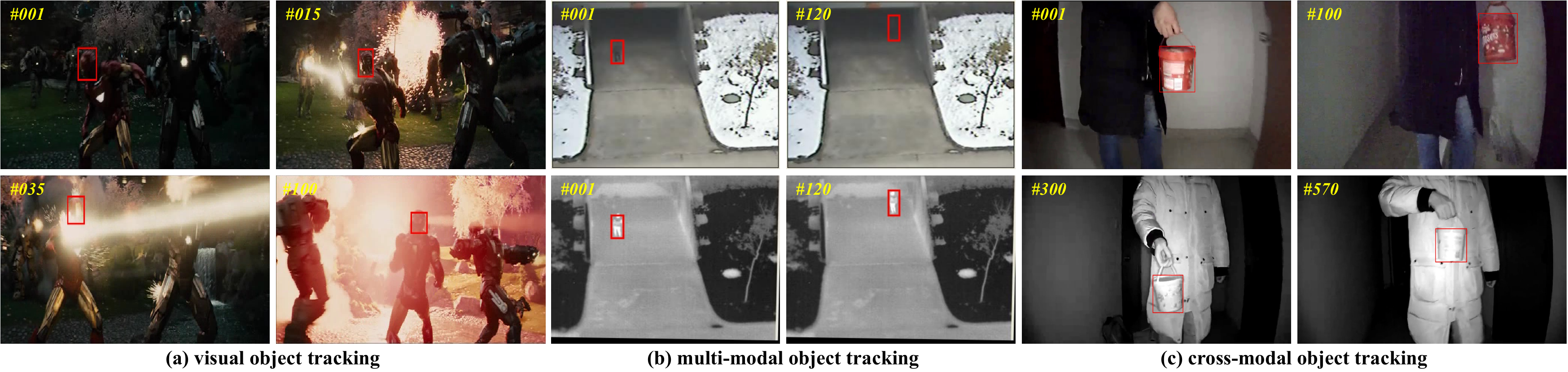}
	%\end{center}
\caption{
Examples of visual object tracking, multi-modal object tracking, and cross-modal object tracking.
% :
% (a) Visual object tracking example from the OTB50 dataset \cite{wu2013online}, where visible video sequences are significantly affected by lighting variations.
% (b) Multi-modal object tracking example from the GTOT dataset \cite{li2016learning}, showcasing video sequences that combine visible and thermal infrared data simultaneously, effectively mitigating the limitations of the visible modality. However, manual spatio-temporal alignment of the two modalities is required.
% (c) Cross-modal object tracking example from our proposed CMOTB dataset, demonstrating adaptive switching between visible and near-infrared modalities in the video sequences based on changes in lighting intensity. This approach leverages the complementary advantages of multi-modal object tracking while avoiding time-consuming and labor-intensive spatio-temporal alignment processes.
}
\label{fig::tracking}
\end{figure*}

Near-infrared (NIR) imaging, commonly used in surveillance cameras\cite{li2020multi}, plays a crucial role by automatically switching between RGB and NIR modalities based on variations in lighting intensity. The NIR wavelength range (700 to 1100 nanometers) provides superior penetration capabilities compared to visible light imaging. It remains unaffected by lighting fluctuations and effectively overcomes the limitations of RGB sources in low-light scenarios. Unlike conventional multi-modal vision systems, this adaptive imaging system dynamically switches modalities based on external lighting conditions, ensuring that only one modality is active at any given time. This eliminates the need for multi-modal data alignment, thereby reducing the time and effort required for the alignment process and addressing the challenges associated with building such a complex multi-modal imaging platform.
Fig.~\ref{fig::tracking} provides visual demonstrations of visual object tracking, multi-modal object tracking, and cross-modal object tracking, illustrating the heterogeneity and distinct visual properties of RGB and NIR modalities. As a result, target objects exhibit significant appearance variations across different modalities, posing a substantial challenge for cross-modal object tracking. Unfortunately, existing tracking methods have not extensively addressed this problem.

In this work, we focus on tackling the challenge of cross-modal object tracking and aim to solve two key questions. Firstly, we aim to develop an algorithm that effectively bridges the appearance gap between RGB and NIR modalities, while also offering flexible integration into different tracking frameworks to ensure robust cross-modal object tracking. Secondly, we aim to create a benchmark dataset of video sequences that can serve as a valuable resource for the advancement of cross-modal object tracking algorithms.

To address the first problem, we propose the \textbf{M}odality-\textbf{A}ware \textbf{F}usion \textbf{Net}work (\textbf{MAFNet}) for adaptive cross-modal object tracking. MAFNet effectively bridges the appearance gap between RGB and NIR modalities by learning modality-specific target representation and modality-aware fusion through an adaptive weighting mechanism based on the modality-aware fusion module (MAFM). One of the key advantages of MAFM is its plug-and-play nature, allowing flexible integration into different tracking frameworks. MAFM consists of two main components: an adaptive weighting module and a modality-specific representation module. The adaptive weighting module predicts fusion weights for the weighting process, enabling dynamic adaptation of the contribution from each modality. This module utilizes a lightweight network with a Sigmoid activation function to handle the challenge of unknown modality switches commonly encountered in real-world scenarios. 

During the training stage, it is trained using modality states as ground truth supervision. This approach accommodates the limitations of commercial sensors, which often cannot acquire the modality state (although it might be available by customizing the sensor). To capture target representations under different modalities, we introduce a modality-specific representation module. This module leverages modality-based Convolutional Neural Networks (CNNs) to extract modality-specific features. By integrating the adaptive weighting module and the modality-specific representation module into the tracking network, MAFNet achieves end-to-end training and adaptively learns modality-specific target representations and modality-aware fusion for training samples in different modalities based on the predicted adaptive fusion weights.

To solve the second problem, we establish a comprehensive benchmark dataset comprising 1000 cross-modal object tracking sequences that cover 61 object categories. The dataset consists of over 799K video frames, with an average sequence length exceeding 799 frames and a maximum length surpassing 2800 frames. To ensure a more comprehensive evaluation of tracker performance in cross-modal scenarios, we divide the dataset into two subsets: the easy set and the hard set. The easy set primarily focuses on common tracking scenarios where each sequence contains modality switches. Conversely, the hard set encompasses more challenging tracking scenarios, such as modality delay caused by the limited adaptability of cross-modal sensors in modality switch scenarios. Each subset is further split into training and testing sets to facilitate separate evaluation. Additionally, we can conduct joint training to obtain more comprehensive evaluation results. Moreover, we introduce three novel challenge attributes specifically designed for cross-modal scenarios. These attributes, namely modality adaptation, modality mutation, and modality delay, represent three typical situations involving modality switches during the tracking process. By evaluating tracker performance on these challenge attributes, we can verify their robustness in cross-modal scenarios.

The main contributions of this paper can be summarized as follows:
\begin{itemize}
\item We introduce a new task called cross-modal object tracking, which is highly challenging and applicable in real-world scenarios.
\item We propose the modality-aware fusion network that effectively addresses appearance differences between different modalities, ensuring robust cross-modal object tracking. Moreover, we integrate this algorithm into two widely-used tracking frameworks to validate its effectiveness.
\item We construct a large-scale unified benchmark dataset for cross-modal object tracking. It offers a strong foundation for advancing cross-modal object tracking research.
\item We conduct extensive experiments to demonstrate the superior performance of our proposed method compared to state-of-the-art trackers.
\end{itemize}

This paper is an extended version of our conference paper \cite{li2022cross} and brings three main improvements. Firstly, we propose a novel method called MAFNet, which enhances the training algorithm compared to the previous MArMOT \cite{li2022cross}. It achieves modality-specific representation learning and modality-aware fusion in an end-to-end training manner, eliminating the need for tedious multi-stage training methods and addressing the issue of sub-optimization. Secondly, we expand the scale of the dataset and consider more realistic challenges in cross-modal scenarios, such as the inadequate adaptation of cross-modal cameras to modality switches, resulting in modality delay. Moreover, we divide the dataset into easy and hard subsets, and for each subset, we have separate training and testing sets. Evaluation can be performed on the easy and hard subsets individually or jointly. Thirdly, MAFNet maintains comparable performance while simplifying the training complexity compared to MArMOT. We observe consistent performance improvements compared to MArMOT on both the easy and hard subsets.

\begin{figure*}[!tbp]
	\centering
    \includegraphics[width=.85\linewidth]{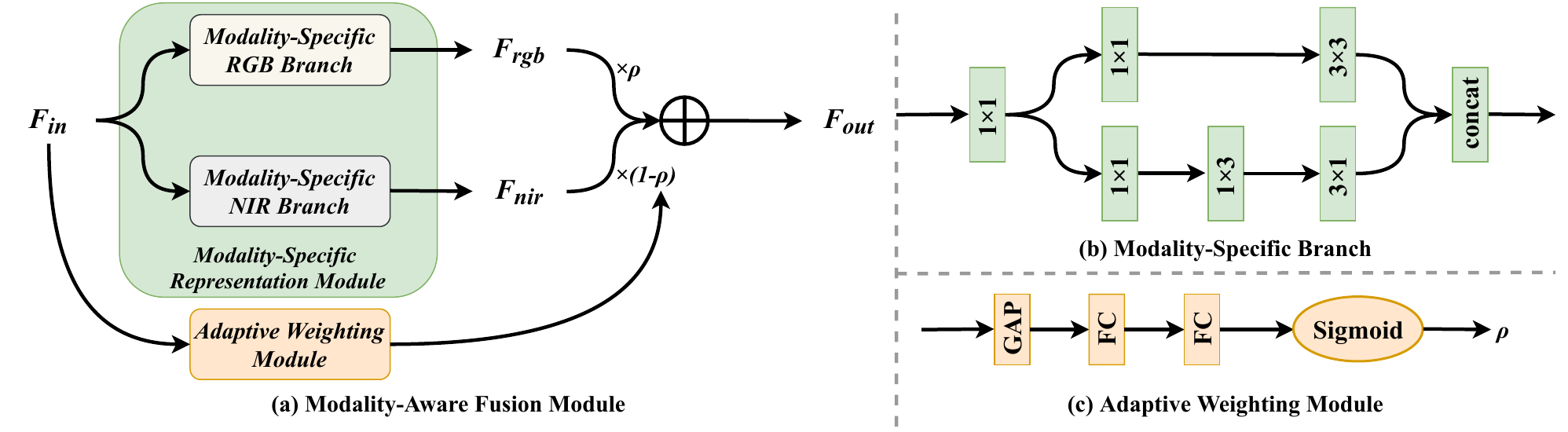}
	%\end{center}
	\caption{Details of the modality-aware fusion module. The BN+ReLU layers after each convolutional layer ($1 \times 1$, $1 \times 3$, $3 \times 1$, $3 \times 3$) and the ReLU layer between different FC layers are omitted for clarity. GAP indicates global average pooling.}
	\label{fig::sub_network}
\end{figure*}

\section{Related Work}
% In this section, we provide a brief overview of relevant research in two areas: visual object tracking and multi-modal object tracking.

\subsection{Visual Object Tracking}
% Visual object tracking is a fundamental research topic in computer vision. Various advanced algorithms have been proposed to address this problem, which 
Existing trackers can be categorized into three typical schemes: classification-based tracking algorithms, matching-based two-stream tracking algorithms, and transformer-based one-stream tracking algorithms.
Classification-based tracking algorithms\cite{nam2016learning, jung2018real} distinguish between foreground and background by training a robust classifier offline. However, due to the adaptability issue of target appearance changes, these methods usually require online updating, resulting in low tracking efficiency.
Matching-based two-stream tracking algorithms\cite{bertinetto2016fully, li2019siamrpn++, zhu2018distractor, zhang2019deeper} treat the tracking task as a template matching problem, using the features of the template image as convolution kernels to convolve the features of the search area. The position with the maximum response is the target to be tracked. These methods are fast but require separate feature extraction and relation modeling, limiting the ability of the network to jointly learn the correlation between the template image and the search area, thus restricting further improvement in tracking performance.
With the widespread application of transformers\cite{vaswani2017attention} in computer vision \cite{dosovitskiy2020vit}, some studies have successfully constructed one-stream tracking frameworks using transformers\cite{ye2022joint, cui2022mixformer}. These frameworks achieve joint feature extraction and relation modeling, simplify the tracking process, and improve tracking accuracy.

Despite the significant performance enhancement achieved by these advanced methods, they rely solely on RGB data, which is limited by inherent drawbacks such as sensitivity to lighting variations. Therefore, achieving robust tracking performance in all-weather and all-day conditions poses a challenge for these methods. In contrast, we exploit the complementary advantages of RGB and NIR data, effectively compensating for the limitations of RGB data.

\subsection{Multi-Modal Object Tracking}
To enhance the adaptability of visual tracking algorithms under varying lighting conditions, researchers have attempted to incorporate data from other modalities to improve their robustness. For instance, Li et al.\cite{li2016learning} utilize thermal infrared data to perceive the thermal radiation information of objects, and Yan et al.\cite{yan2021depthtrack} employ depth data to perceive distance information, while Wang et al.\cite{wang2023visevent} introduce event data to perceive motion information. Additionally, several algorithms have been proposed to effectively utilize multiple modalities, including pixel-level fusion\cite{chan2013fusing, zhang2022visible}, feature-level fusion\cite{wang2023visevent, liu2023quality, xiao2022attribute}, and decision-level fusion\cite{feng2020learning}.

Although the integration of multi-modal data 
% and the development of multi-modal algorithms 
have significantly improved tracking performance under harsh lighting conditions, precise pixel-level alignment of multi-modal data is necessary to fully leverage their complementary advantages, which is a time-consuming and labor-intensive process. Moreover, constructing multi-modal imaging platforms is complex and expensive, typically costing thousands to tens of thousands of dollars. In contrast, our cross-modal tracking method can effectively exploit the complementary advantages of multi-modal data while avoiding the tedious multi-modal data alignment process. Additionally, cross-modal cameras are easily obtainable and inexpensive in surveillance scenarios, costing only a few dozen dollars.

\section{Proposed Method}
In this section, we introduce our proposed method, the Modality-Aware Fusion Module (MAFM), for adaptive cross-modal object tracking. MAFM aims to bridge the appearance gap between RGB and NIR modalities and enable modality-aware fusion of target representations. It consists of two main components: an adaptive weighting module and a modality-specific representation module. These components work together to effectively fuse information from both modalities.
Next, we describe the integration of MAFM with two typical tracking frameworks. This integration allows MAFM to seamlessly combine with different trackers, leveraging their existing infrastructure and capabilities to enhance cross-modal object tracking performance.
Then, we delve into the details of the end-to-end training algorithm employed by MAFM. Through joint optimization, both the adaptive weighting module, the modality-specific representation module, and the tracking framework are trained simultaneously.
Finally, we provide a detailed explanation of the online tracking process.

\subsection{Modality-Aware Fusion Module}
In the task of cross-modal object tracking, the RGB and NIR modalities exhibit significant differences in their visual properties, posing a challenge for cross-modal object tracking. To address this issue, we propose the Modality-Aware Fusion Module (MAFM) for modality-specific target representation learning and modality-aware fusion, to alleviate the appearance gap between RGB and NIR modalities during the tracking process, as shown in Fig.~\ref{fig::sub_network}(a).

MAFM employs an adaptive weighting mechanism to dynamically fuse information and capture specific target representations from different modalities. It consists of an adaptive weighting module and a modality-specific representation module. The adaptive weighting module predicts fusion weights for the adaptive weighting process, allowing dynamic adjustment of the contribution from each modality. The modality-specific representation module takes into account the visual properties of RGB and NIR modalities and captures discriminative features specific to each modality. It uses modality-based Convolutional Neural Networks (CNNs) to extract modality-specific target representation, thereby enhancing the fusion of information from RGB and NIR modalities. Importantly, MAFM offers great flexibility as it can effortlessly integrate into diverse tracking frameworks.

\subsubsection{Modality-Specific Representation Module}
NIR imaging switches to RGB based on light intensity, leading to significant variations in target appearance. Therefore, it is necessary to model target representations under different modalities. In particular, we design a modality-specific representation module, as illustrated in Fig.~\ref{fig::sub_network}(a). It is implemented using modality-based CNN with two branches, each responsible for modeling modality-specific target representations. This design simplifies the modeling of cross-modal target representation with distinct appearance differences.

For efficiency and effectiveness, each branch utilizes an inception-like network~\cite{szegedy2017inception}. In each modality-specific branch, the first $1\times1$ convolutional layer captures the representation of modality-specific information. It is then split into two streams using two additional $1\times1$ convolutional layers with half the channels, reducing the dimension of the input feature and computational complexity. Subsequently, two types of $3\times3$ convolutions are employed to enhance the network's adaptability in learning target features of different scales. Finally, the outputs are concatenated to form the modality-specific target representation. Details of the modality-specific branch are depicted in Fig.~\ref{fig::sub_network}(b).

\subsubsection{Adaptive Weighting Module}
In the training stage, the modality state is known. However, in the tracking process, it is often unavailable in commercial sensors (although it might be available by customizing the sensor). Therefore, handling scenarios with unknown modality switches is essential.

To address this problem, we design an adaptive weighting module that predicts fusion weights to determine the contribution of each modality in the adaptive weighting process. We use a lightweight network with a Sigmoid activation function to handle unknown modality switches commonly encountered in real-world scenarios. During the training stage, the adaptive weighting module is trained using modality states as ground truth supervision.
\begin{figure*}[!tbp]
	\centering
    \includegraphics[width=.85\linewidth]{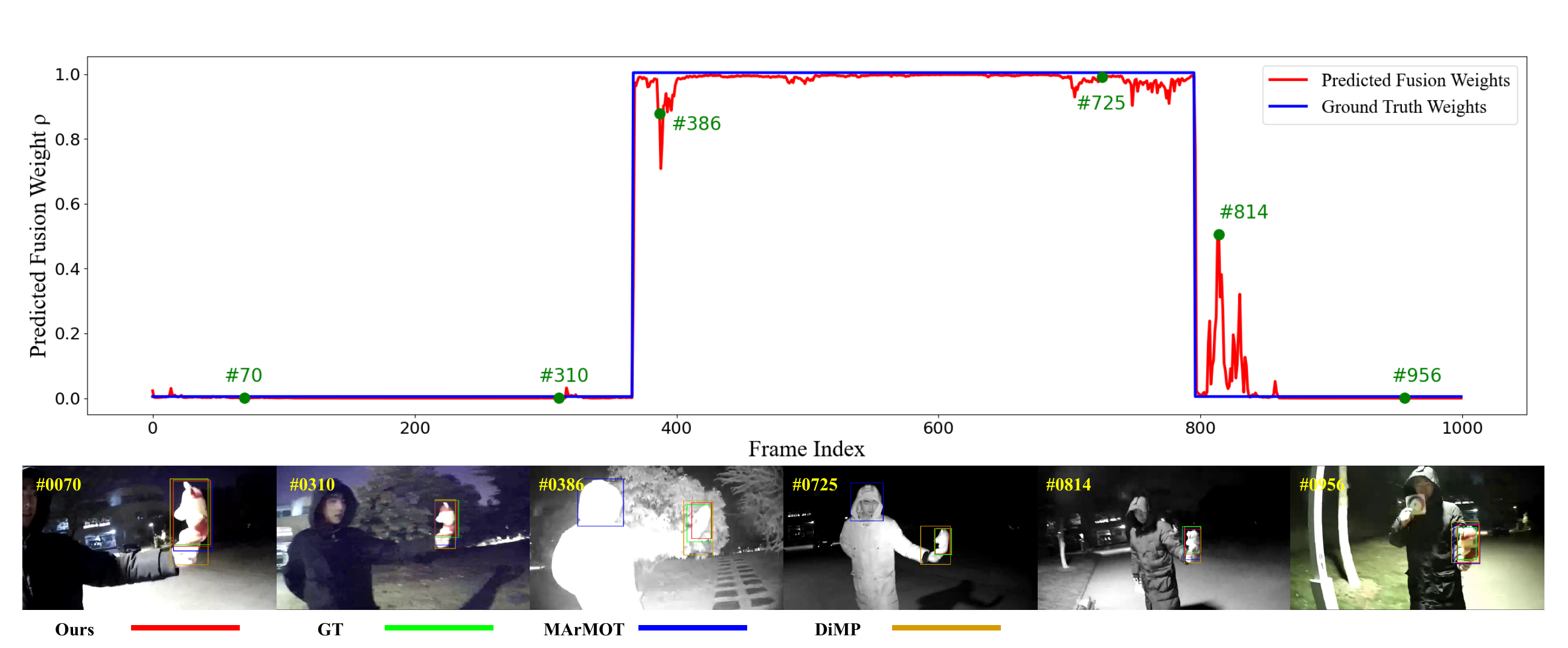}
	%\end{center}
	\caption{
 Effectiveness of fusion weight predictions and representative tracking results in a video sequence. The curves demonstrate the accurate fusion weight predictions of our MAFNet, enhancing the robustness compared to the baseline tracker DiMP~\cite{bhat2019learning} and MArMOT~\cite{li2022cross}.
	}
	\label{fig::motivation}
\end{figure*}

As shown in Fig.~\ref{fig::sub_network}, our adaptive weighting module consists of a Global Average Pooling (GAP) layer and two Fully Connected (FC) layers, followed by a Sigmoid function. This predicts a normalized fusion weight that is assigned to different modality-specific branches for the adaptive weighting fusion. The fusion process is defined by the equation:

\begin{equation}
	F_{out} = \rho \times F_{rgb} + (1 - \rho) \times F_{nir}
	\label{eq:equation1}
\end{equation}
where $\rho$ is the predicted fusion weight, $F_{rgb}$ and $F_{nir}$ are the modality-specific features corresponding to different modalities, and $F_{out}$ is the modality-aware fusion feature. By predicting fusion weights based on the observed data, MAFM can adaptively adjust the contribution of each modality even in the absence of explicit modality switch signals. During the training stage, the supervision of the adaptive weighting module is the ground truth modality state, and the loss function is the binary cross-entropy loss.

In Fig.~\ref{fig::motivation}, we present the results of adaptive weight prediction for a typical sequence, comparing them with the tracking results of the baseline tracker DiMP~\cite{bhat2019learning} and MArMOT~\cite{li2022cross}. It is evident that our method demonstrates high accuracy in both fusion weight prediction and tracking results, thereby validating the effectiveness of our approach.

\begin{figure*}[!htbp]  %[]
	\centering
	\includegraphics[width=.85\linewidth]{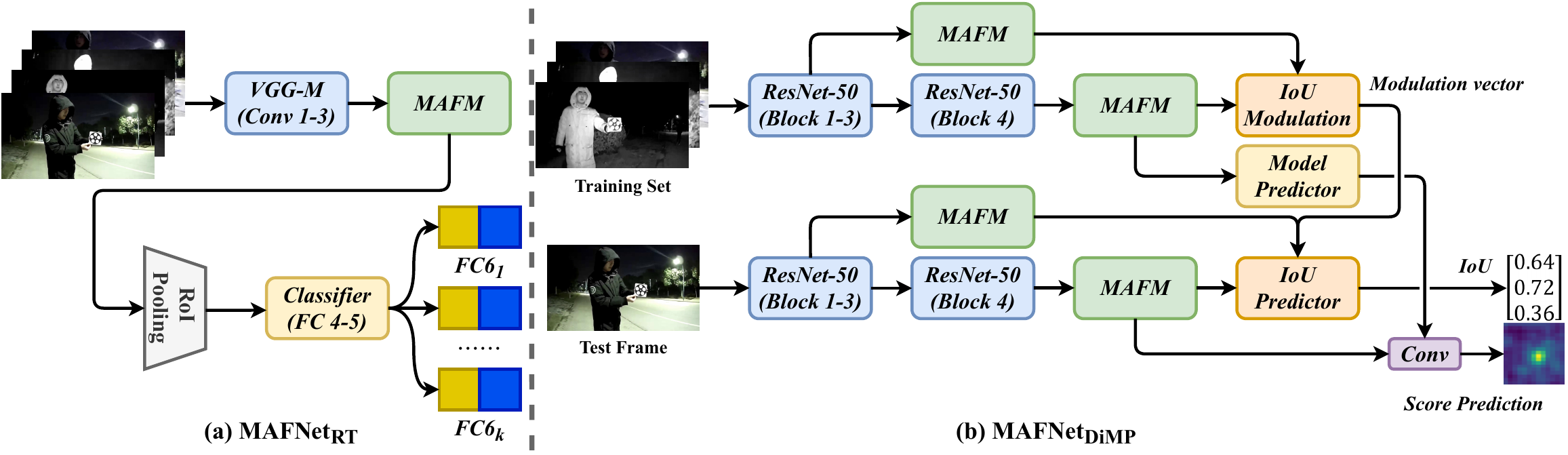}
	\caption{Visualization of the tracking architectures with MAFM. (a) and (b) show the detailed structures of MAFM combined with RT-MDNet and DiMP, respectively.}
	\label{fig::overall_network}
\end{figure*}

\subsection{Integration with Tracking Frameworks}
Existing tracking frameworks typically use one-stream backbone networks to extract features, making it challenging to handle significant appearance differences caused by modality switches in cross-modal tracking. To address this issue, we propose adding MAFM behind the feature extraction network of the existing tracker.

Specifically, we embed the proposed plug-and-play MAFM into two tracking frameworks: RT-MDNet~\cite{jung2018real} and DiMP~\cite{bhat2019learning}, named MAFNet$\bf_{RT}$ and MAFNet$\bf_{DiMP}$, respectively, to verify the effectiveness and generalization of MAFM. The overall tracking frameworks are shown in Fig.~\ref{fig::overall_network}.
For each tracking framework, we first use the backbone network to extract deep feature representations of the target. Then, we embed MAFM to bridge the appearance gap between RGB and NIR modalities by learning modality-aware fusion target representations through the proposed adaptive weighting mechanism. Finally, the modality-aware fusion feature is sent to the classification branch and regression branch to perform target localization.

For RT-MDNet, it starts with several convolutional layers borrowed from VGG-M~\cite{chatfield2014return}, which capture common low-level information across modalities. Therefore, we insert MAFM after the last convolutional layer to learn modality-aware target representations fusion. This design also reduces computational complexity since the feature map's size in the last layer is the smallest, as shown in Fig.~\ref{fig::overall_network}(a).
For DiMP, the backbone network connects the IoU predictor and model predictor simultaneously, using different layer features. Thus, we need to insert MAFM modules after different layers for these two predictors, as shown in Fig.~\ref{fig::overall_network}(b). The consideration for reducing computational complexity is the same as the design in RT-MDNet.
\begin{figure*}[!tbp]
	\centering
    \includegraphics[width=.9\linewidth]{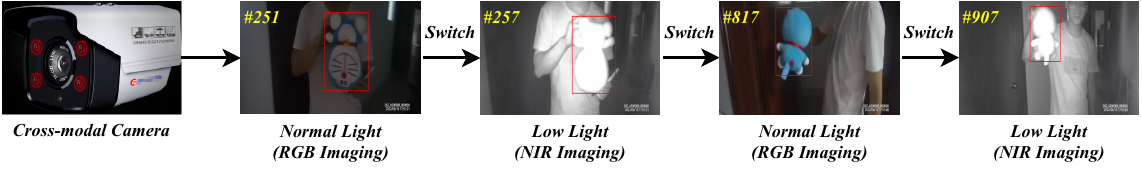}
	%\end{center}
\caption{
Illustration of heterogeneous properties between RGB and NIR modalities. The cross-modal camera changes RGB imaging to NIR when the light intensity becomes low from normal, and vice versa.}
\label{fig::plotv3}
\end{figure*}

\begin{figure}[!tbp]
	\centering
    \includegraphics[width=.8\linewidth]{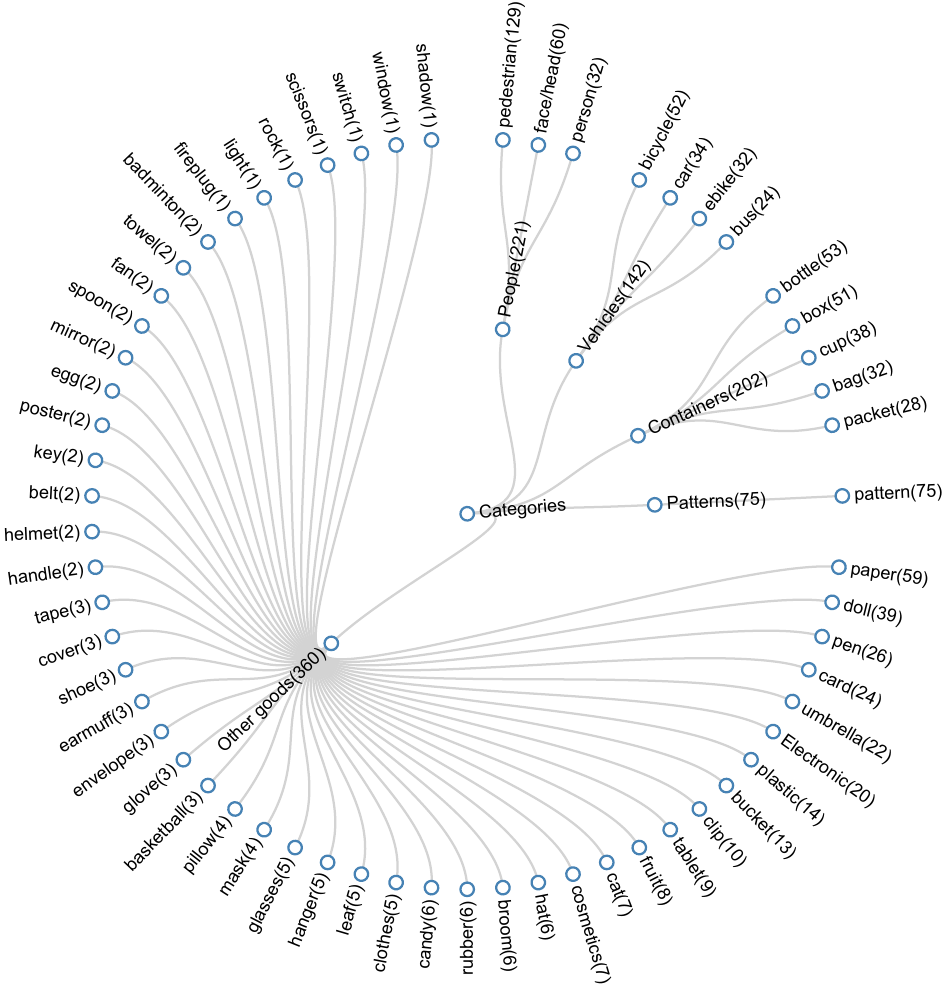}
	%\end{center}
\caption{
Distribution of object categories in the CMOTB dataset.}
\label{fig::category}
\end{figure}
\subsection{End-to-End Training Algorithm}
Researchers have explored various approaches to address the challenge of single-branch networks struggling with diverse complex scenarios in tracking tasks. These approaches include decoupling the learning of target representations based on different attributes and utilizing separate branches to model these attributes. For example, ANT \cite{qi2019learning} and CAT \cite{li2020challenge} focus on learning attribute-specific representations, while ADRNet \cite{zhang2021learning} and APFNet \cite{xiao2022attribute} focus on integrating multi-branch representations. In cross-modal object tracking, MArMOT \cite{li2022cross} decouples modality-specific representation learning. However, these methods often involve multi-stage learning for adaptive multi-branch representation fusion, making the training process complex and lacking end-to-end optimization, which may lead to suboptimal solutions.

To achieve end-to-end training of the proposed modality-aware fusion module and tracking framework, we use the modality weighting scheme to perform effective fusion of different modalities. This scheme is based on the adaptive weighting operation, which is differentiable and can be integrated into the end-to-end learning framework. By jointly optimizing the network parameters, MAFNet effectively fuses information from RGB and NIR modalities, achieving modality-specific target representation learning and modality-aware fusion.

\subsubsection{Loss Function}
% {\flushleft\bf Loss Function.}
The loss function $L$ is a weighted sum of the loss for the adaptive weighting module $L_{\text{weight}}$ and the loss for the tracking framework $L_{\text{track}}$:
\begin{equation}
	L = \alpha \times L_{\text{weight}} + \beta \times L_{\text{track}}
	\label{eq:equation3}
\end{equation}
where $\alpha$ and $\beta$ represent the weights of the different losses. In our experiments, we set $\alpha$ and $\beta$ to 1. Through the optimization of the adaptive weighting module, modality-specific representation module, and the tracking framework in an end-to-end training manner, the adaptive weighting weights can adapt to variations in input cross-modal data and maintain feature discrimination when modality switches occur.

\subsubsection{Training Details}
% {\flushleft\bf Training Details.}
We initialize our model with the parameters of the backbone network pretrained on large-scale datasets such as ImageNet\cite{deng2009imagenet} and GOT-10K\cite{huang2019got}. %, LaSOT\cite{fan2019lasot}, COCO\cite{lin2014microsoft}, and TrackingNet\cite{muller2018trackingnet}
The whole model is then trained in an end-to-end manner. In our setting, the learning rate of the network parameters (excluding MAFM) is set to one-tenth of the default learning rate of the tracking framework. The learning rates of the modality-specific representation module and the adaptive weighting module are set to 5e-5 and 5e-4, respectively. The number of iterations is the same as that of the basic tracking framework.

\subsection{Online Tracking}
The online tracking procedures and parameter configurations of our trackers closely resemble those of the fundamental tracking framework. However, a notable distinction lies in the utilization of deep features extracted by the backbone networks (VGG-M\cite{chatfield2014return} in MAFNet$_{RT}$ and ResNet50\cite{he2016deep} in MAFNet$_{DiMP}$). These features are then fed into our proposed MAFM for modality-aware fusion, effectively mitigating the appearance discrepancies present in target representations captured across different modalities. The resulting outputs from MAFM serve as inputs to both the classifiers (FC4-FC6 in MAFNet$_{RT}$ and the model predictor in MAFNet$_{DiMP}$) and the regressor (IoU predictor module in MAFNet$_{DiMP}$). For a comprehensive understanding of the tracking processes, please refer to Fig.~\ref{fig::overall_network} for a detailed visual representation.

\section{CMOTB Benchmark Dataset}
% In the field of cross-modal object tracking, researchers require large-scale datasets for training advanced trackers and evaluating different tracking algorithms. The CMOTB addresses this need by providing a comprehensive and diverse \textbf{C}ross-\textbf{M}odal \textbf{O}bject \textbf{T}racking \textbf{B}enchmark. In this section, we provide a detailed analysis and introduction to CMOTB.
\subsection{Data Collection and Annotation}
\subsubsection{Large-scale Collection}
% {\flushleft \bf Large-scale Collection.}
To overcome the scarcity of cross-modal video data in visual object tracking research, we create the CMOTB dataset. Our primary objective is to offer a large-scale and diverse dataset that accurately represents real-world scenarios and challenges. To achieve this, we use hand-held cameras to capture video data from various scenes with different background complexities. Unlike traditional visual object tracking data, we also consider variations in light intensity that trigger modality switches during data creation. Additionally, we carefully select environmental conditions to simulate real-world applications such as visual surveillance, intelligent transportation, and autonomous driving systems. Fig.~\ref{fig::plotv3} shows a typical example of the CMOTB dataset, where the imaging switches between RGB and NIR modalities several times.
Specifically, we collect a total of 1000 cross-modal image sequences, comprising over 481K frames in total. The average video length exceeds 735 frames, covering 61 different object categories. The distribution of these categories is illustrated in Fig.~\ref{fig::category}.
\begin{figure}[!t]
	\centering
    \includegraphics[width=\linewidth]{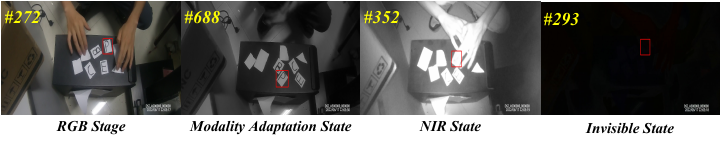}
	%\end{center}
\caption{
Examples of frame-level modality states in the CMOTB dataset.}
\label{fig::fourstate}
\end{figure}
\begin{figure*}[!tbp]
	\centering
    \includegraphics[width=.9\linewidth]{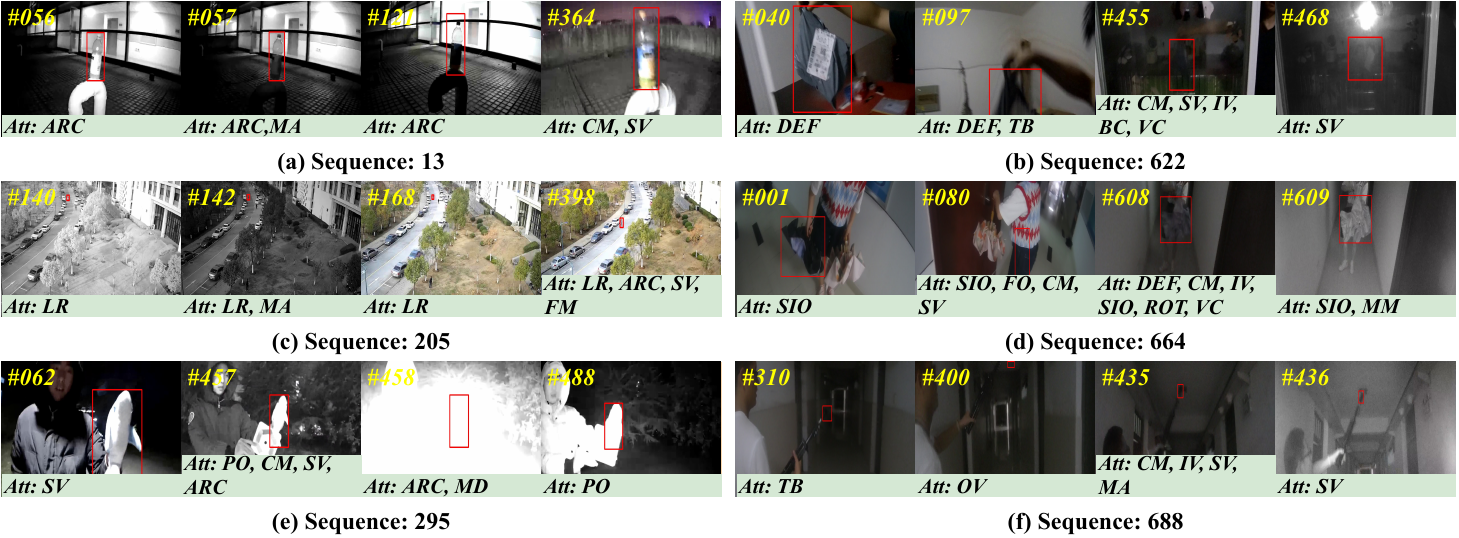}
	%\end{center}
\caption{
Sample frames from the CMOTB dataset. Attributes are shown at the bottom.}
\label{fig::showseq}
\end{figure*}
\subsubsection{High-quality Dense Annotation}
% {\flushleft \bf High-quality Dense Annotation.}
To provide accurate annotations, we represent target object states using minimum bounding boxes that capture position and scale. Due to the time-consuming nature of the labeling process, we have developed an auxiliary labeling tool based on ViTBAT~\cite{biresaw2016vitbat}. This tool allows for manual or semi-automatic labeling of target object states through a simple and user-friendly interface, improving efficiency. The generated bounding boxes are accurate in most situations. However, when the target object undergoes drastic appearance variations, the generated bounding boxes may not be entirely accurate. In such cases, we manually adjust the bounding boxes with great care.

To ensure high-quality annotations, we have trained a team of four professional annotators to adhere to consistent annotation standards. Additionally, professional checkers perform a frame-by-frame inspection to prevent incorrect and inaccurate labels. In scenarios where modality switches occur and objects become temporarily invisible, we maintain the ground truth unchanged for the target object until it becomes visible again.

In addition to target object state annotations, we provide frame-level modality state annotations, including four states: RGB state, modality adaptation state, NIR state, and invisible state, to support the learning of modality states. Examples of these four states can be seen in Fig.~\ref{fig::fourstate}.

\begin{table}[!t]
\caption{Descriptions of attributes in the CMOTB dataset.}
\centering
\begin{tabular}{lp{200pt}}
\hline
  & Definition \\
\hline
\textbf{FO} & Full Occlusion - Target fully occluded. \\
\textbf{PO} & Partial Occlusion - Target partially occluded. \\
\textbf{DEF} & Deformation - Target undergoes non-rigid movement. \\
\textbf{SV} & Scale Variation - Ratio of current to initial bounding box outside $\tau \in [0.5, 2]$. \\
\textbf{ROT} & Rotation - Target object rotates. \\
\textbf{FM} & Fast Motion - Motion larger than bounding box size. \\
\textbf{CM} & Camera Motion - Abrupt camera movement. \\
\textbf{IV} & Illumination Variation - Changes in target illumination. \\
\textbf{TB} & Target Blur - Blurry target appearance. \\
\textbf{OV} & Out-of-View - Target completely missing. \\
\textbf{BC} & Background Clustering - Similar color/texture to target in the background. \\
\textbf{SIO} & Similar Interferential Object - Objects visually resembling the target. \\
\textbf{LR} & Low Resolution - Bounding box area less than 400. \\
\textbf{ARC} & Aspect Ratio Change - Significant change in bounding box aspect ratio outside [0.5, 2]. \\
\textbf{VC} & Viewpoint Change - Change in target viewpoint. \\
\textbf{MA} & Modality Adaptation - Slow modality switch with pseudo intermediate modality. \\
\textbf{MM} & Modality Mutation - Rapid modality switch without pseudo intermediate modality. \\
\textbf{MD} & Modality Delay - Delayed modality switch due to inadequate adaptation. \\
\hline
\end{tabular}
\label{tab::attribute}
\end{table}

\subsection{Attributes}
Existing multi-modal tracking datasets
% , such as RGBT tracking~\cite{li2016learning, li2020challenge}, RGBD tracking~\cite{song2013tracking, yan2021depthtrack}, and RGBE tracking~\cite{wang2023visevent}, 
include two-modal data in each frame, whereas our dataset contains only one modality per frame but may exhibit modality switches. This is the major difference from existing multi-modal visual object tracking datasets.

Considering the modality switch, we introduce three new attributes called modality adaptation, modality mutation, and modality delay, representing three typical situations involving modality switches during the tracking process. By evaluating tracker performance on these challenging attributes, we verify their robustness in cross-modal scenarios. Specifically, modality adaptation refers to a slow modality switch leading to the appearance of a pseudo intermediate modality. Modality mutation refers to a rapid modality switch without going through a pseudo intermediate modality. Modality delay refers to a delay in modality switch caused by inadequate adaptation of the device to environmental changes, such as frames that are completely white or black.

To enable attribute-based performance analysis of trackers, we annotate each sequence with 18 attributes, including Full Occlusion (FO), Partial Occlusion (PO), Deformation (DEF), Scale Variation (SV), Rotation (ROT), Fast Motion (FM), Camera Motion (CM), Illumination Variation (IV), Target Blur (TB), Out-of-View (OV), Background Clustering (BC), Similar Interferential Object (SIO), Low Resolution (LR), Aspect Ratio Change (ARC), Viewpoint Change (VC), Modality Adaptation (MA), Modality Mutation (MM), and Modality Delay (MD). We provide detailed definitions for each attribute in Table~\ref{tab::attribute} and present visual examples of some sequences containing these 18 attributes in Fig.~\ref{fig::showseq} for a more intuitive demonstration.
\begin{figure*}[!htbp]  %[]
	\centering
	\includegraphics[width=.8\linewidth]{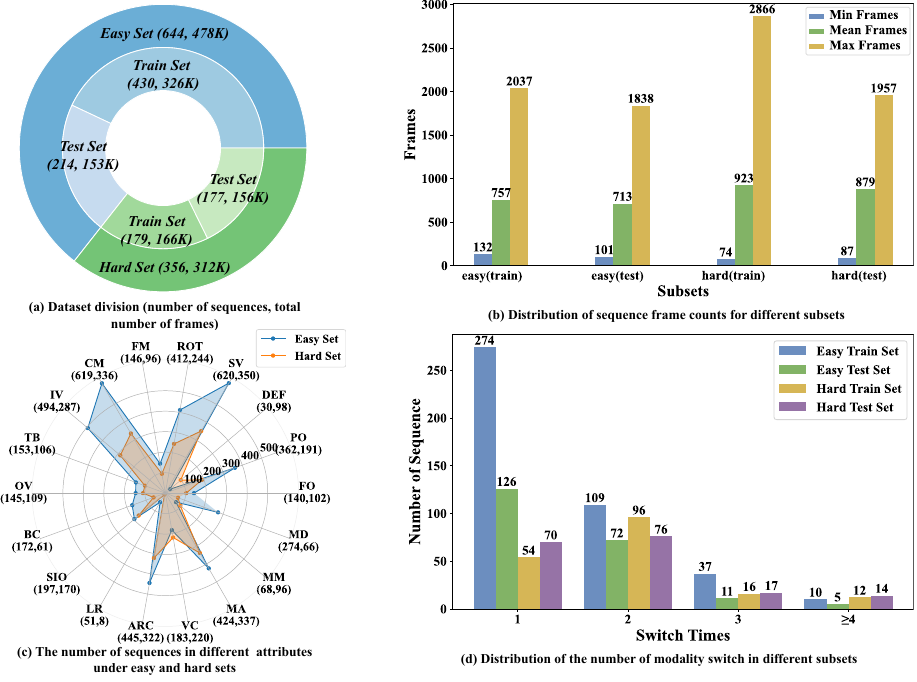}
\caption{Information statistics of the CMOTB dataset.}
\label{fig::static_all3}
\end{figure*}
\subsection{Statistics}
The CMOTB dataset consists of 1000 video sequences, covering a wide range of challenges encountered in real-world scenarios. To ensure a comprehensive evaluation of tracker performance in cross-modal scenarios, we divide the dataset into two subsets: the easy set and the hard set. The easy set primarily focuses on common tracking scenarios where each sequence contains modality switches, and it consists of 644 video sequences. The hard set encompasses more challenging tracking scenarios, such as modality delay caused by the limited adaptability of cross-modal sensors in modality switch scenarios, and it consists of 356 video sequences. Since there are no other cross-modal object tracking datasets available, each subset is further split into training and test sets to facilitate separate evaluation. Additionally, joint training can be conducted to obtain more comprehensive evaluation results. The specific division of each subset of CMOTB is shown in Fig.~\ref{fig::static_all3}(a), and the distribution of sequence frame counts for each subset is shown in Fig.~\ref{fig::static_all3}(b).

Modality switch refers to the change in imaging modality caused by variations in light intensity. In such scenarios, the appearance of the target object usually varies significantly, making it challenging for trackers. The number of modality switches in a sequence is a key factor that affects the performance of trackers. Therefore, we record the number of switch times during data creation and report the data distribution of switch times in Fig.~\ref{fig::static_all3}(d). Additionally, we also summarize the distribution of different attributes in Fig.~\ref{fig::static_all3}(c).
\begin{table*}[!tbp]
\caption{The PR, NPR, and SR scores (\%) of various trackers on different subsets. The best and second-best results are highlighted in $\color{red} red$ and $\color{blue} blue$ colors, respectively.}
\centering
\begin{tabular}{@{}lcccccccccccc@{}}
\toprule
\multirow{2}{*}{Methods} & \multirow{2}{*}{Publication} & \multirow{2}{*}{Classification} & \multicolumn{3}{c}{Easy Set} & \multicolumn{3}{c}{Hard Set} & \multicolumn{2}{c}{Joint Set} &  & \multirow{2}{*}{FPS} \\ \cmidrule(lr){4-12}
 &  &  & PR & NPR & SR & PR & NPR & SR & PR & NPR & SR &  \\ \midrule
MAFNet$\bf_{DiMP}$ & ours & \multirow{16}{*}{\begin{tabular}[c]{@{}c@{}}Matching-based \\ Two-stream Methods\end{tabular}} & \color{red} 74.4 & \color{red}77.8 & \color{red}63.8 & 32.2 & \color{red}53.6 & \color{red}47.0 & \color{red}55.1 & \color{red}66.4 & \color{red}56.1 & 36 \\
MArMOT$\bf_{DiMP}$ & AAAI 2022 &  & \color{blue}73.9 & \color{blue}77.4 & \color{blue}63.6 & 30.6 & \color{blue}52.6 & 46.3 & \color{blue} 53.8 & \color{blue}64.5 & \color{blue}54.6 & 33 \\
ToMP & CVPR 2022 &  & 59.4 & 62.0 & 52.7 & 25.6 & 42.2 & 40.1 & 44.1 & 53.0 & 47.0 & 25 \\
TrDiMP & CVPR 2021 &  & 60.2 & 62.5 & 53.9 & 24.7 & 41.1 & 40.0 & 44.1 & 52.8 & 47.6 & 26 \\
TransT & CVPR 2021 &  & 58.0 & 59.6 & 51.0 & 29.6 & 43.9 & 41.2 & 45.2 & 52.5 & 46.5 & 50 \\
Ocean & ECCV 2020 &  & 47.7 & 50.8 & 42.5 & 17.0 & 31.5 & 30.2 & 33.8 & 42.0 & 36.9 & 42 \\
SiamBAN & CVPR 2020 &  & 55.6 & 58.3 & 47.7 & 23.8 & 40.7 & 35.7 & 41.2 & 50.3 & 42.3 & 29 \\
TACT & ACCV 2020 &  & 40.4 & 43.3 & 38.7 & 15.6 & 32.7 & 31.9 & 29.2 & 38.5 & 35.6 & 29 \\
DiMP & ICCV 2019 &  & 58.1 & 61.3 & 51.3 & 22.7 & 38.3 & 36.0 & 42.1 & 50.9 & 44.4 & 41 \\
ATOM & CVPR 2019 &  & 56.2 & 58.7 & 48.3 & 21.7 & 35.8 & 31.4 & 40.6 & 48.4 & 40.7 & 28 \\
SiamRPN++ & CVPR 2019 &  & 54.6 & 57.1 & 47.1 & 22.6 & 38.9 & 35.1 & 40.1 & 48.9 & 41.6 & 21 \\
SiamMask & CVPR 2019 &  & 54.3 & 56.4 & 45.7 & 22.0 & 37.2 & 32.0 & 39.7 & 47.7 & 39.5 & 41 \\
SiamDW & CVPR 2019 &  & 40.3 & 44.7 & 35.2 & 12.6 & 26.3 & 23.2 & 27.8 & 36.4 & 29.8 & 22 \\
GradNet & ICCV 2019 &  & 44.6 & 46.5 & 36.5 & 13.7 & 26.5 & 22.5 & 30.6 & 37.5 & 30.2 & 68 \\
DaSiamRPN & ECCV 2018 &  & 51.1 & 54.5 & 43.4 & 9.0 & 24.5 & 21.0 & 32.1 & 40.9 & 33.3 & \color{blue}144 \\
SiamFC & ECCVW 2016 &  & 43.3 & 45.2 & 37.3 & 11.6 & 20.9 & 18.5 & 28.9 & 34.2 & 28.8 & 44 \\ \midrule
MAFNet$\bf_{RT}$ & ours & \multirow{5}{*}{\begin{tabular}[c]{@{}c@{}}Classification-based \\ Methods\end{tabular}} & 53.0 & 56.2 & 43.7 & 16.2 & 29.9 & 23.5 & 35.6 & 43.7 & 33.8 & 23 \\ %\cmidrule(r){1-2} \cmidrule(l){4-13} 
MArMOT$\bf_{RT}$ & AAAI 2022 &  & 48.9 & 52.0 & 39.8 & 15.5 & 28.5 & 22.7 & 34.5 & 42.6 & 32.4 & 23 \\
RT-MDNet & ECCV 2018 &  & 41.8 & 44.4 & 35.3 & 13.6 & 25.3 & 20 & 29.1 & 35.8 & 28.4 & 29 \\
VITAL & CVPR 2018 &  & 29.7 & 31.4 & 26.3 & 16.1 & 31.3 & 27.1 & 23.6 & 31.3 & 26.7 & 0.3 \\
MDNet & CVPR 2016 &  & 51.6 & 54.2 & 42.7 & 16.5 & 33.5 & 29.2 & 35.7 & 44.8 & 36.6 & 1 \\ \midrule
ROMTrack & ICCV 2023 & \multirow{9}{*}{\begin{tabular}[c]{@{}c@{}}Transformer-based \\ One-stream Methods\end{tabular}} & 55.2 & 56.8 & 49.5 & 31.6 & 45.2 & 42.0 & 44.5 & 51.6 & 46.1 & 62 \\
%\cmidrule(r){1-2} \cmidrule(l){4-13} 
% ROMTrack & ICCV 2023 & & 55.2 & 56.8 & 49.5 & 31.6 & 45.2 & 42.0 & 44.5 & 51.6 & 46.1 & 62\\
GRM & CVPR 2023 & & 50.5 & 51.8 & 46.0 & 29.8 & 42.8 & 40.4 & 41.1 & 47.7 & 43.4 & 45\\
MixFormerV2 & NeurIPS 2023 & & 52.2 & 54.1 & 47.8 & 28.5 & 41.7 & 39.8 & 41.5 & 48.5 & 44.2 & \color{red}165\\
DropTrack & CVPR 2023 & & 60.6 & 61.7 & 53.8 & \color{red} 35.9 & 50.5 & \color{blue} 46.8 & 49.4 & 56.7 & 50.6 & 30\\
SeqTrack & CVPR 2023 & & 57.9 & 60.1 & 51.0 & 31.8 & 47.3 & 43.3 & 46.1 & 54.3 & 47.5 & 40\\
ARTrack & CVPR 2023 & & 59.7 & 60.7 & 53.1 & \color{blue}34.3& 47.8 & 44.8 & 48.2 & 54.9 & 49.4 & 26\\
AiATrack & ECCV 2022 & & 62.1 & 64.2 & 54.9 & 31.7 & 48.6 & 45.6 & 48.3 & 57.2 & 50.7 & 38 \\ 
OSTrack & ECCV 2022 &  & 51.0 & 52.2 & 46.2 & 28.3 & 41.6 & 39.6 & 40.7 & 47.4 & 43.2 & 93 \\
 %\cmidrule(r){1-2} \cmidrule(l){4-13} 
Stark & ICCV 2021 &  & 56.0 & 57.8 & 50.4 & 26.8 & 41.5 & 40.2 & 42.8 & 50.4 & 45.8 & 30 \\ %\midrule
\bottomrule
\end{tabular}
\label{tab::results}
\end{table*}
\subsection{Evaluation Metrics}
In our experiments, we utilize the one-pass evaluation (OPE) protocol to evaluate the performance of all trackers. We employ three widely used evaluation metrics in the tracking field~\cite{muller2018trackingnet}: precision rate (PR), normalized precision rate (NPR), and success rate (SR).

\begin{itemize}
    \item PR (Precision Rate): The precision rate quantifies the percentage of frames where the center distance between the predicted position and the ground truth position is smaller than a predefined threshold, denoted as $\tau$. For our experiments, we set $\tau$ to 20.
    \item NPR (Normalized Precision Rate): To account for variations in the size of the tracked object across different sequences, we introduce the normalized precision rate as an additional evaluation metric. NPR facilitates fair comparisons by normalizing the precision rate (PR) based on the average size of the ground truth bounding box. This normalization helps alleviate any bias towards trackers that perform better on either larger or smaller objects, enabling a more comprehensive assessment of precision.
    \item SR (Success Rate): In addition to precision, we evaluate the success rate to measure tracker effectiveness. SR assesses the ratio of tracked frames, determined by the Intersection-over-Union (IoU) between the tracking result and the ground truth. We construct a success plot (SP) by varying the overlap thresholds to visualize tracker performance at different levels of overlap. The success rate (SR) is then computed as the area under the curve of the SP, providing an overall measure of tracking success.
\end{itemize}

% To ensure a comprehensive evaluation, we consider both the overall performance across all sequences and the attribute-based performance for attribute-specific sequences. This approach allows us to assess tracker performance in different scenarios and challenges.

\subsection{Discussion}
\subsubsection{Differences From Relevant Tasks}
% {\bf \flushleft Differences From Relevant Tasks.}
Our task of cross-modal object tracking differs from the task of multi-modal visual object tracking in several aspects. Existing work usually introduces thermal infrared, depth, or event data to achieve multi-modal visual object tracking.
% , known as RGBT tracking~\cite{li2016learning, li2020challenge}, RGBD tracking~\cite{song2013tracking, yan2021depthtrack}, and RGBE tracking~\cite{wang2023visevent}. 
Comparing our task with multi-modal visual object tracking, we have identified the following differences and advantages. First, our task is more practical as many visual cameras are already equipped with NIR imaging, whereas other multi-modal imaging platforms require two cameras. Second, our task is more cost-effective as other multi-modal imaging platforms are typically expensive, while our task only relies on surveillance cameras, eliminating these limitations. Finally, the multi-modal data in our task do not have any alignment errors compared to other multi-modal visual tracking tasks that involve two cameras and require alignment across different modalities. In contrast, our imaging system includes only one camera, which can switch between RGB and NIR modalities.
\subsubsection{Acquisition of Modality Switch Signals}
% {\bf \flushleft Acquisition of Modality Switch Signals.}
To the best of our knowledge, commercial sensors currently do not provide signals for modality switches. However, it might be possible to obtain such signals by customizing the sensor. Therefore, studying how to handle scenarios with unknown modality switches is essential.

% \begin{figure}[t]  %[]
% 	\centering
% 	\includegraphics[width=.9\linewidth]{staticv4.pdf}
% \caption{CMOTB Dataset Division and Frame Count Distribution.
% }
% \label{fig::staticv2}
% \end{figure}

% \begin{figure}[t]  %[]
% 	\centering
% 	\includegraphics[width=\linewidth]{switch.pdf}
% \caption{Distribution of the number of modality switch in different subsets.
% }
% \label{fig::modalityswitch}
% \end{figure}

% \begin{figure}[t]  %[]
% 	\centering
% 	\includegraphics[width=\linewidth]{radar_chart.pdf}
% \caption{Chart of Attributes, the chart showcases the number of sequences in the easy set, the hard set, and the all set in different attributes.
% }
% \label{fig::attribute}
% \end{figure}

\section{Experiments}
% In this section, we evaluate the performance of our algorithm by comparing it with state-of-the-art trackers on three different cross-modal tracking subsets: the easy set, hard set, and joint set. These subsets will help validate the effectiveness of our proposed approach. Additionally, we analyze the efficiency of each major component in the algorithm. 
Our tracker is implemented using PyTorch on a PC computer equipped with an Intel Xeon Gold 5220R CPU and two NVIDIA A100 GPUs.

\subsection{Comparison with State-of-the-art Trackers}
\subsubsection{Evaluated Algorithms}
We evaluate 28 advanced and representative trackers on our benchmark. These trackers encompass the mainstream tracking algorithms developed between 2016 and 2023. The evaluated trackers in our study can be divided into three types.
The first type consists of 15 matching-based two-stream methods, including MArMOT$\bf_{DiMP}$~\cite{li2022cross}, ToMP~\cite{mayer2022transforming}, TrDiMP~\cite{wang2021transformer}, TransT~\cite{chen2021transformer}, Ocean~\cite{zhang2020ocean}, SiamBAN~\cite{chen2020siamese}, TACT~\cite{choi2020visual}, DiMP~\cite{bhat2019learning}, ATOM~\cite{danelljan2019atom}, SiamRPN++~\cite{li2019siamrpn++}, SiamMask~\cite{wang2019fast}, SiamDW~\cite{zhang2019deeper}, GradNet~\cite{li2019gradnet}, DaSiamRPN~\cite{zhu2018distractor}, and SiamFC~\cite{bertinetto2016fully}.
The second type includes 4 classification-based methods, namely MArMOT$\bf_{RT}$~\cite{li2022cross}, RT-MDNet~\cite{jung2018real}\footnotemark
\footnotetext{In our approach, to adapt to our experimental platform, we replace the RoI Align layer in RT-MDNet with the Precise RoI Pooling~\cite{jiang2018acquisition} layer.}, VITAL~\cite{song2018vital}, and MDNet~\cite{nam2016learning}.
The third type consists of 9 transformer-based one-stream methods, which are ROMTrack~\cite{cai2023robust}, GRM~\cite{gao2023generalized}, MixFormerV2~\cite{cui2023mixformerv2}, DropTrack~\cite{wu2023dropmae}, SeqTrack~\cite{chen2023seqtrack}, ARTrack~\cite{wei2023autoregressive}, AiAtrack~\cite{gao2022aiatrack}, OSTrack~\cite{ye2022joint}, and Stark~\cite{yan2021learning}.
It is important to note that we evaluate the performance of these algorithms on our testing set using the provided models from the authors.

\begin{figure*}[!htbp]
    \centering
    \includegraphics[width=.9\linewidth]{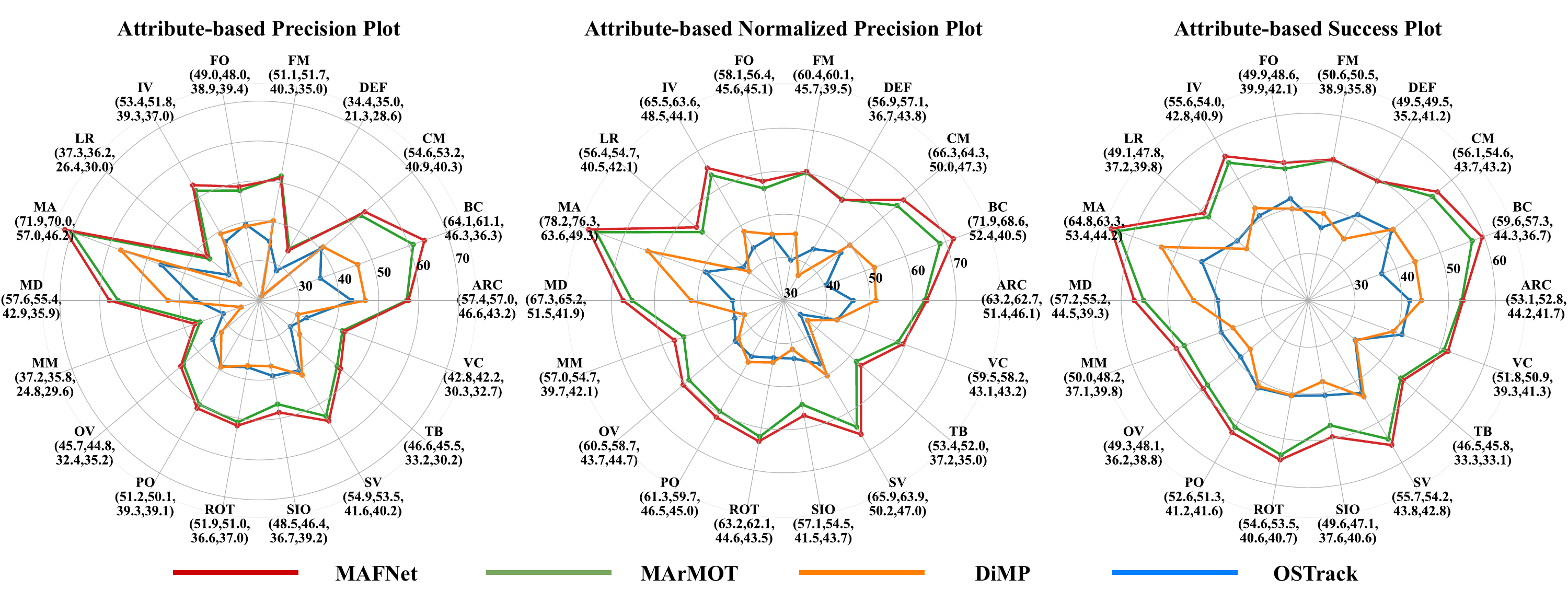}
    \caption{Comparison based on attributes in the joint set.}
    \label{fig::rate2}
\end{figure*}

\subsubsection{Overall Performance}
% Table~\ref{tab::results} presents a comprehensive comparison of the performance of our algorithm with 28 other competing algorithms on different subsets of CMOTB. 
% The table clearly demonstrates that our method outperforms the other algorithms across all subsets, showcasing its significant superiority in terms of three metrics: PR, NPR, and SR.
{\bf \flushleft Comparison with Matching-based Two-stream Trackers.}
Matching-based two-stream tracking algorithms have long been the mainstream solution in the field of visual object tracking, as they possess efficient inference speed and excellent tracking accuracy. Typically, these algorithms employ an offline-trained regressor to perform the tracking process. However, these algorithms face challenges in cross-modal object tracking scenarios due to significant appearance differences between different modalities, as shown in Table~\ref{tab::results}. To validate the effectiveness of our approach, we integrate MAFM into a typical tracking framework, DiMP, creating MAFNet$\bf_{DiMP}$.

In particular, MAFNet$\bf_{DiMP}$ significantly outperforms the baseline DiMP, demonstrating its effectiveness. In the easy set, MAFNet$\bf_{DiMP}$ achieves improvements of 16.3\%, 16.5\%, and 12.5\% under PR, NPR, and SR metrics, respectively. In the hard set, the improvements are 9.5\%, 15.3\%, and 11.0\%, and in the joint set, they are 13.0\%, 15.5\%, and 11.7\%.

Furthermore, compared to MArMOT$\bf_{DiMP}$, our method consistently achieves performance improvements across all subsets. Specifically, in the easy set, MAFNet$\bf_{DiMP}$ shows improvements of 0.5\%, 0.4\%, and 0.2\% under PR, NPR, and SR metrics, respectively. In the hard set, the improvements are 1.6\%, 1.0\%, and 0.7\%, and in the joint set, they are 1.3\%, 1.9\%, and 1.5\%. It is important to note that both our method and MArMOT$\bf_{DiMP}$ employ modality-aware fusion techniques. However, MArMOT$\bf_{DiMP}$ requires a time-consuming and resource-intensive multi-stage training process, whereas our framework adopts an end-to-end training approach, significantly simplifying the training complexity while improving performance.

{\bf \flushleft Comparison with Classification-based Trackers.} 
Classification-based tracking methods typically utilize offline learning of a generic classifier and achieve discrimination between the target and background through online fine-tuning. To validate the generalization of our approach, we incorporate MAFM into a typical classification-based tracking framework, RT-MDNet, creating MAFNet$\bf_{RT}$. Similarly, MAFNet$\bf_{RT}$ demonstrates significant enhancements compared to the baseline RT-MDNet tracker. In the easy set, it achieves improvements of 11.2\%, 11.8\%, and 8.4\% under PR, NPR, and SR metrics, respectively. In the hard set, the improvements are 2.6\%, 4.6\%, and 3.5\%, and in the joint set, they are 6.5\%, 7.9\%, and 6.4\%.

Furthermore, compared to MArMOT$\bf_{RT}$, our method consistently achieves performance improvements across all subsets. Specifically, in the easy set, our method shows improvements of 4.1\%, 4.2\%, and 3.9\% on the PR, NPR, and SR metrics, respectively. In the hard set, our method demonstrates improvements of 0.7\%, 1.4\%, and 1.2\%, and in the joint set, our method exhibits improvements of 1.1\%, 1.1\%, and 1.4\%.

{\bf \flushleft Comparison with Transformer-based One-stream Trackers.}
Transformer-based one-stream tracking methods have gained increasing attention in recent years due to their significant performance advantages over other types of tracking methods, achieved by unifying the processes of feature extraction and relational modeling. However, the application of our MAFM on these frameworks presents certain challenges as it currently relies solely on CNN. In the future, we aim to explore the design of suitable structures for transformer-based trackers to address the issue of significant cross-modal appearance variations. Nonetheless, we have already conducted performance comparisons with some classical transformer-based trackers.

For example, compared to OSTrack, our MAFNet$\bf_{DiMP}$ demonstrates performance improvements in the easy set with increases of 1.2\%, 3.0\%, and 1.3\% for the PR, NPR, and SR metrics, respectively. In the hard set, the improvements are 0.2\%, 3.3\%, and 2.1\%, and in the joint set, the improvements are 4.2\%, 1.0\%, and 1.1\%.

These experimental results provide strong evidence for the simplicity and effectiveness of our method in cross-modal scenarios. Additionally, these results illuminate the challenges faced by existing algorithms in dealing with significant appearance variations caused by modality switches in cross-modal tracking scenarios. In the subsequent paper, unless otherwise specified, when referring to MAFNet and MArMOT, we are referring to the variants based on DiMP.

\subsubsection{Attribute-based Performance}
To validate the robustness of trackers in cross-modal tracking scenarios, we evaluate the performance of different trackers across 18 attributes, as shown in Fig.~\ref{fig::rate2}. For clarity, we only display the results of the four typical trackers. From the figure, it is evident that our method achieves significant improvements across all attributes. Particularly, our method demonstrates a more pronounced advantage in tackling challenges related to modality switches, such as MM, MA, and MD attributes. These results strongly demonstrate the effectiveness of our method in addressing cross-modal tracking scenarios.

\begin{figure}[!tbp]  %[]
	\centering
	\includegraphics[width=.9\linewidth]{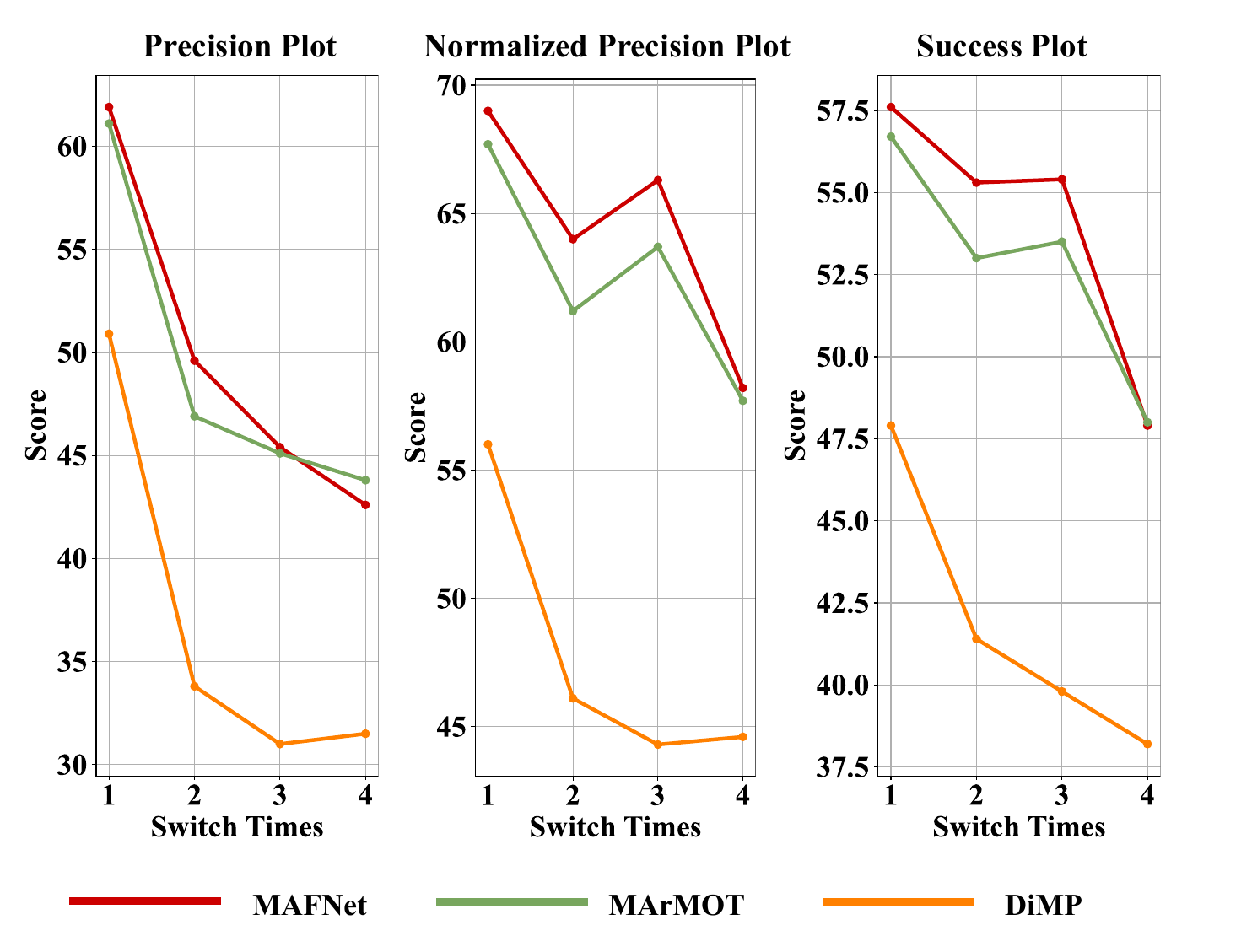}
\caption{Comparison of the 3 trackers under varying switch times in the joint set.}
\label{fig::switchperformance}
\end{figure}

\begin{figure*}[!htbp]  %[]
	\centering
	\includegraphics[width=.86\linewidth]{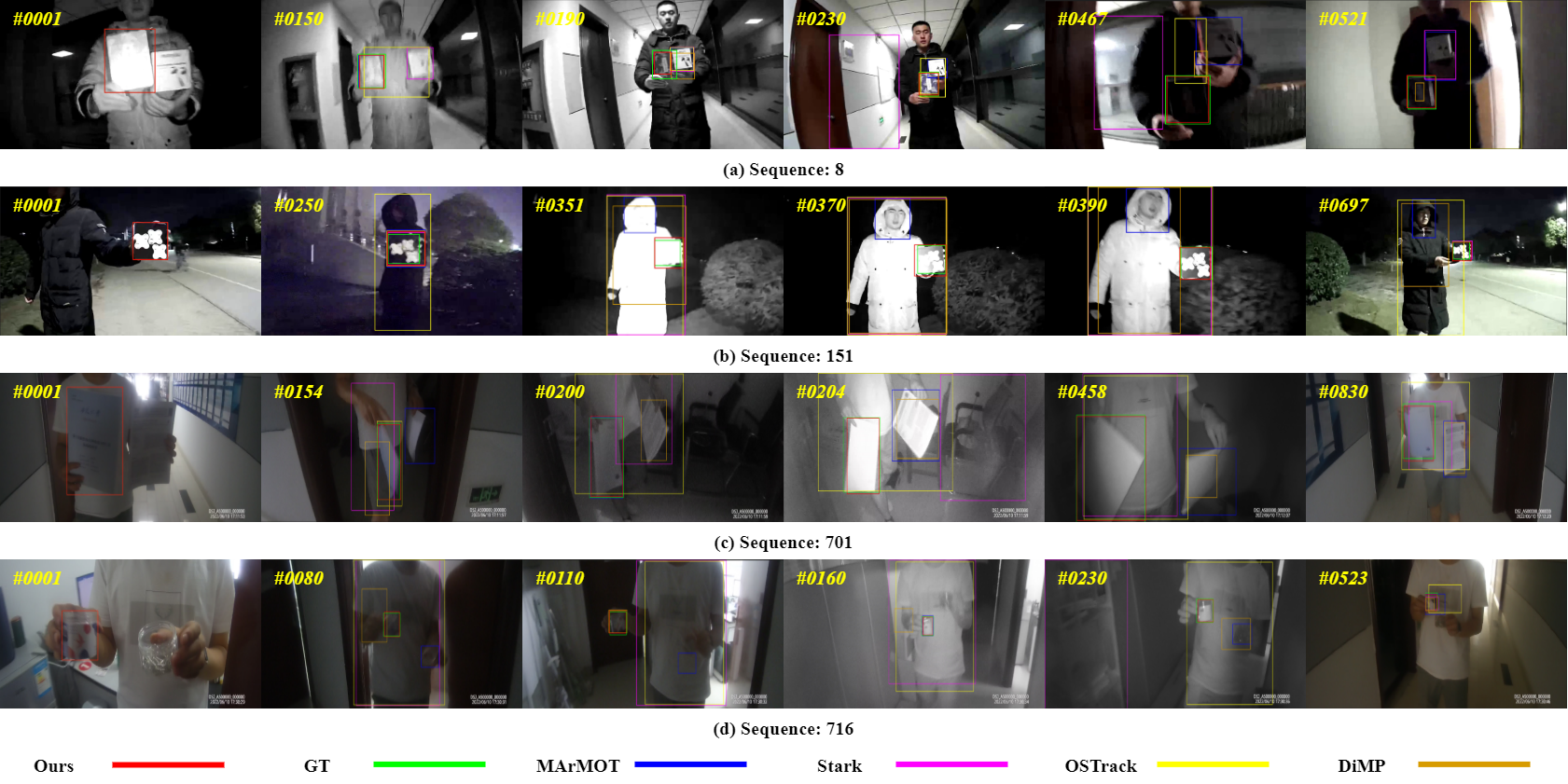}
\caption{Visual comparison on four representative sequences.}
\label{fig::Qualitative}
\end{figure*}
\subsubsection{Impact of Switch Times}
% {\bf \flushleft Impact of Switch Times.}
Modality switch is a crucial factor that affects the performance of cross-modal trackers. To validate the impact of modality switch times on tracker performance, we present the performance results under different switch times in Fig.~\ref{fig::switchperformance}. For clarity, we only display the results of the three typical trackers. From the figure, it can be observed that as the modality switch times increase, the performance of the baseline tracker, DiMP, exhibits a consistent downward trend. This emphasizes the greater challenges introduced by modality switch in cross-modal tracking.
However, it is noteworthy that our method, MAFNet, as well as MArMOT, exhibit improved performance in NPR and SR metrics when the switch times are three compared to the case when it is two. This indicates the advantage of our methods in addressing the challenges posed by modality switches.
\begin{table}[!tbp]
\caption{The PR, NPR, and SR scores (\%) of various trackers on different subsets. * indicates that the tracker is re-trained using the joint training set. The best and second-best results are highlighted in $\color{red} red$ and $\color{blue} blue$ colors, respectively.}
\resizebox{.5\textwidth}{!}{
\begin{tabular}{llllllllll}
\hline
\multicolumn{1}{c}{{Methods}} &
  \multicolumn{3}{c}{Easy Set} &
  \multicolumn{3}{c}{Hard Set} &
  \multicolumn{3}{c}{Joint Set} \\ \cline{2-10} 
\multicolumn{1}{c}{} &
  \multicolumn{1}{c}{PR} &
  \multicolumn{1}{c}{NPR} &
  \multicolumn{1}{c}{SR} &
  \multicolumn{1}{c}{PR} &
  \multicolumn{1}{c}{NPR} &
  \multicolumn{1}{c}{SR} &
  \multicolumn{1}{c}{PR} &
  \multicolumn{1}{c}{NPR} &
  \multicolumn{1}{c}{SR} \\ \hline
MDNet  & 51.6 & 54.2 & 42.7 & 16.5 & 33.5 & 29.2 & 35.7 & 44.8 & 36.6  \\
MDNet* & 58.2 & 61.8 & 48.1 & 17.5 & 35.3 & 30.1 & 41.0 & 51.8 & 41.1  \\
RT-MDNet & 41.8 & 44.4 & 35.3 & 13.6 & 25.3 & 20.0 & 29.1 & 35.8 & 28.4 \\
RT-MDNet* & 45.6 & 49.3 & 37.2 & 16.2 & 29.0 & 22.5 & 32.3 & 40.1 & 30.5 \\
DiMP & 58.1 & 61.3 & 51.3 & 22.7 & 38.3 & 36.0 & 42.1 & 50.9 & 44.4      \\
DiMP* & 70.4 & 73.8 & 60.0 & 30.8 & 50.4 & 45.1 & 52.5 & 63.2 & 53.3     \\ 
TrDiMP & 60.2 & 62.5 & 53.9 & 24.7 & 41.1 & 40.0 & 44.1 & 52.8 & 47.6      \\
TrDiMP* & 66.8 & 69.9 & 58.3 & 29.1 & 46.2 & 41.2 & 49.7 & 59.2 & 50.6      \\ 
TransT & 58.0 & 59.6 & 51.0 & 29.6 & 43.9 & 41.2 & 45.2 & 52.5 & 46.5      \\
TransT* & 68.9 & 72.0 & 59.1 & 30.4 & 49.9 & 43.4 & 51.5 & 62.0 & 52.0      \\ 
OSTrack & 51.0 & 52.2 & 46.2 & 28.3 & 41.6 & 39.6 & 40.7 & 47.4 & 43.2\\
OSTrack* & \color{blue}74.3 & \color{blue}76.8 & \color{blue}63.3 & \color{red}35.9 & \color{blue}52.7 & \color{blue}46.0 & \color{red}56.9 & \color{blue}65.9 & \color{blue}55.5\\
\hline
MAFNet$\bf_{DiMP}$ & \color{red}74.4 & \color{red}77.8 & \color{red}63.8 & \color{blue}32.2 & \color{red}53.6 & \color{red}47.0 & \color{blue}55.1 & \color{red}66.4 & \color{red}56.1 \\
MAFNet$\bf_{RT}$ & 53.0 & 56.2 & 43.7 & 16.2 & 29.9 & 23.5 & 35.6 & 43.7 & 33.8\\
\hline
\end{tabular}
}
\label{tab::retrain}
\end{table}
\subsubsection{Training Dataset Validation}
% {\bf \flushleft Training Dataset Validation.}
To demonstrate the necessity of constructing a large-scale cross-modal dataset, we retrain six trackers (RT-MDNet, MDNet, DiMP, TrDiMP, TransT and OSTrack) using the training data from CMOTB. The experimental results, presented in Table \ref{tab::retrain}, clearly indicate significant performance improvements for all trackers after retraining. This finding highlights the importance of our proposed cross-modal dataset in driving advancements in cross-modal object tracking research.

Furthermore, by integrating MAFNet into the DiMP and RT-MDNet frameworks, we observe further performance improvements. Specifically, we achieve enhancements of 2.6\%/3.2\%/2.8\% and 3.3\%/3.6\%/3.3\% on the three metrics, respectively, for both tracking frameworks under the joint set. This provides strong evidence for the effectiveness of our proposed approach in addressing cross-modal tracking tasks.

\subsection{Qualitative Analysis}
To provide an intuitive demonstration of the performance of our MAFNet, we present qualitative results in Fig.~\ref{fig::Qualitative}. It showcases visual comparisons of our MAFNet with several state-of-the-art trackers on four representative sequences. These sequences encompass challenging scenarios with appearance variations, offering a rigorous evaluation of tracking performance. Our visual results prominently highlight the remarkable ability of MAFNet in handling appearance differences caused by cross-modal variations. Across all sequences, our method consistently outperforms other trackers after modality changes, demonstrating the robustness and adaptability of our approach in tracking objects across different modalities.

For instance, in Sequence 8, a sudden modality switch occurs at frame 230, which poses a significant challenge for most trackers. However, our MAFNet adeptly adapts to the new modality and accurately localizes the target. This showcases the superior performance of our method in handling modality switches.

Furthermore, in Sequence 151, while many trackers can successfully track the target in the RGB modality, their performance significantly deteriorates in the NIR modality. In contrast, our method maintains precise and reliable tracking in both modalities consistently. Additionally, in Sequences 701 and 716, our method stably tracks the target even amidst frequent modality switches, whereas other trackers struggle to achieve consistent tracking results. These cases demonstrate the capability of our algorithm to effectively adapt to cross-modal appearance variations, thereby validating the effectiveness of modality-aware fusion.

% \begin{table}[!htbp]
% \caption{Tracking results on the CMOTB joint test set using the proposed end-to-end training method and multi-stage training method, respectively.}
% \centering
% \begin{tabular}{@{}cccccc@{}}
% \toprule
% {Method} &
%   \multicolumn{3}{c}{Joint Set} &
%   {\begin{tabular}[c]{@{}c@{}}Total\\ Training\\ Time\end{tabular}} &
%   {\begin{tabular}[c]{@{}c@{}}Total\\ Training\\ Params\end{tabular}} \\ \cmidrule(lr){2-4}
%             & PR & NPR & SR &  &  \\ \midrule
% MAFNet          & 55.1 & 66.4 & 56.1 & 22.5 hours & 58.5 M \\
% MArMOT          & 53.8 & 65.4 & 54.6 & 59.2 hours & 75.95 M \\
% MArMOT$_{E2E}$ &  52.2  & 63.0 & 52.8 & 24.6 hours &  58.4 M\\ \bottomrule
% \end{tabular}
% \label{tab::e2e}
% \end{table}

\begin{table}[]
\caption{Tracking results on the CMOTB joint test set using the proposed end-to-end training method and the multi-stage training method, respectively.}
\centering
\begin{tabular}{@{}cccccc@{}}
\toprule
\multirow{2}{*}{Method} & \multicolumn{3}{c}{Joint Set} & \multirow{2}{*}{\begin{tabular}[c]{@{}c@{}}Total\\ Training Time\(\downarrow\) \end{tabular}} & \multirow{2}{*}{\begin{tabular}[c]{@{}c@{}}Total\\ Training Params\(\downarrow\) \end{tabular}} \\ \cmidrule(lr){2-4}
 & PR & NPR & SR &  &  \\ \midrule
MAFNet & 55.1 & 66.4 & 56.1 & 22.5 hours & 58.5 M \\
MArMOT & 53.8 & 65.4 & 54.6 & 59.2 hours & 75.95 M \\ %\midrule
MArMOT$\bf_{E2E}$ & 52.2 & 63.0 & 52.8 & 24.6 hours & 58.4 M \\ \bottomrule
\end{tabular}
\label{tab::e2e}
\end{table}

\subsection{Effectiveness of End-to-End Training}
In this paper, we face the challenging task of achieving end-to-end learning for modality-aware fusion. To address this problem, we ingeniously design an adaptive weighting mechanism to enable modality-specific representation learning and modality-aware fusion. Similar to our approach, MArMOT also aims to achieve modality-specific representation modeling and adaptive integration. However, they require a multi-stage training scheme to achieve their goal as they are unable to accomplish it within an end-to-end framework.

To validate the effectiveness and efficiency of our end-to-end training approach, we compare it with MArMOT in terms of training costs (total training time and total trainable parameters) and performance. Please refer to Table~\ref{tab::e2e} for detailed results. From the table, we observe that our end-to-end training strategy achieves higher performance compared to the multi-stage training scheme employed in MArMOT, with only one-third of the training time and fewer trainable parameters required. Conversely, if MArMOT tries to adopt an end-to-end training approach (referred to as MArMOT$\bf_{E2E}$), the results indicate that under end-to-end training, MArMOT$\bf_{E2E}$ fails to effectively learn modality-specific representations, leading to a noticeable decline in tracking performance.

\section{Conclusion}
In this paper, we present a large-scale benchmark dataset with high-quality dense bounding box annotations for cross-modal object tracking. The dataset is divided into easy and hard subsets based on the complexity of the tracking scenarios. Additionally, we propose a simple yet effective method for cross-modal object tracking based on modality-aware fusion network. The method is validated in two typical tracking frameworks, showcasing its generalization. By employing an end-to-end learning approach, we jointly train the tracking network, adaptive weighting module, and modality-specific representation module. Through extensive experimentation on the cross-modal tracking dataset, our method surpasses the current state-of-the-art trackers in terms of performance. By releasing this dataset, we believe it will have a positive impact on the research and development of cross-modal object tracking. In the future, we will continue to explore more efficient tracking algorithms to address the challenges of cross-modal tracking.

\bibliographystyle{IEEEtran}
\bibliography{mafnet}

\end{document}